# Memory Matching Networks for Genomic Sequence Classification

**Jack Lanchantin, Ritambhara Singh, & Yanjun Qi**
Department of Computer Science
University of Virginia
Charlottesville, VA 22903, USA
`{jjl5sw,rs3zz,yq2h}@cs.virginia.edu`

## Abstract

When analyzing the genome, researchers have discovered that proteins bind to DNA based on certain patterns on the DNA sequence known as "motifs". However, it is difficult to manually construct motifs due to their complexity. Recently, external learned memory models have proven to be effective methods for reasoning over inputs and supporting sets. In this work, we present memory matching networks (MMN) for classifying DNA sequences as protein binding sites. Our model learns a memory bank of encoded motifs, which are dynamic memory modules, and then matches a new test sequence to each of the motifs to classify the sequence as a binding or non binding site.

## 1 Introduction

In genomics, Transcription Factors (TFs) are proteins which bind to certain locations on a DNA sequence and in turn control gene regulation. Thus, predicting the Transcription Factor Binding sites (TFBSs), or locations where TFs bind on the genome, is particularly useful for understanding genomic processes and improving human health. Biologists have discovered that the binding of a TF is triggered by local sequential patterns within TFBSs, known as "motifs" (Stormo, 2000). As a result, researchers tried to predict TFBSs by manually constructing motifs using position weight matrices (PWMs) which best represented the positive binding sites, and then compared the test sequence to the PWMs to see if there is a close match. However, it is difficult to find accurate PWMs due to the wide variety of TFBS sequences. Additionally, a TFBS may be influenced by a combination of different motifs. In this work, we attempt to create a memory bank of fully learnable motifs, which we can then read from to compare with and classify a new test sequence.

On the TFBS task, researchers initially used PWM-matching approaches (Stormo, 2000), which were manually created motifs for comparison. This was later outperformed by a convolutional neural network (CNN) model which could learn PWM-like filters (Alipanahi et al., 2015). In our model, we use a learned memory bank of PWM-like matrices which we directly use for matching against the test sequence. We run our model on a baseline TFBS dataset and compare against PWM, CNN, and LSTM approaches.

Recently, deep learning models which use external memory have emerged as promising methods for many tasks (Graves et al., 2014; Sukhbaatar et al., 2015; Vinyals et al., 2015; Miller et al., 2016). These studies focus on reasoning using external memory based on the input in order to produce some target. In our work, we focus on creating a model which uses comparison over a learned memory bank of motifs for classification. Our work is closely related to the "matching network" (MN) model by Vinyals et al. (2016), where they train a differentiable nearest neighbor model to find the closest matching image from a support set on a new unseen image. We modify the MN model with a dynamic memory support set for the TFBS classification task. The key difference is that our memory matching network model is for a general classification setting (i.e. not one-shot) where we seek to instead *learn* the support set (memory units), which remains constant for every new test classification. We believe that the dynamically learned memory can function similarly to the motifs that traditional manually constructed methods used. Our preliminary models show that the learned memory can find the binding patterns better than traditional deep learning methods, and also provide similar explanatory patterns that are predominantly used in bioinformatics.

## 2 MEMORY MATCHING NETWORK MODEL FOR CLASSIFICATION

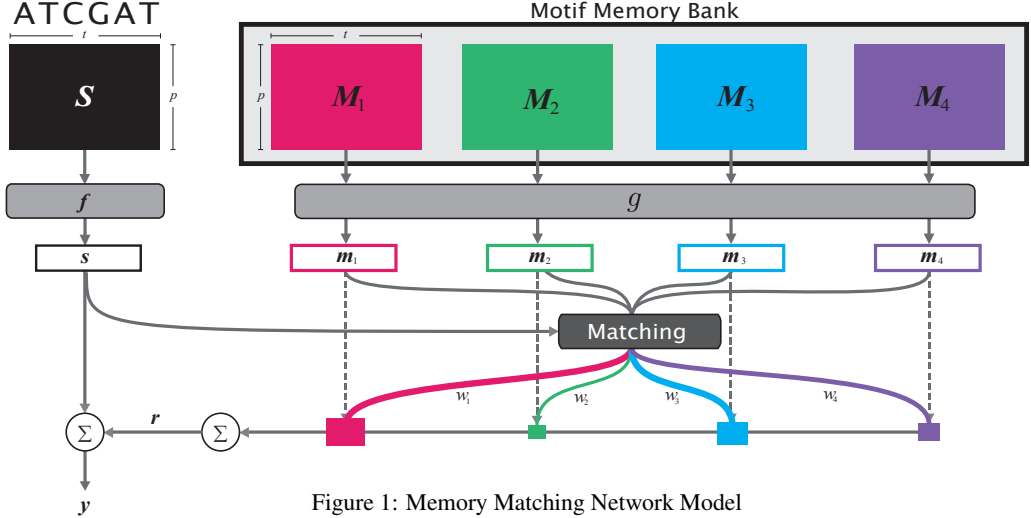

Figure 1: Memory Matching Network Model

Given a DNA sequence $S$ (composed of characters A,C,G,T) of length $t$ and a transcription factor protein of interest, we want to classify $S$ as a positive or negative binding site for this particular TF (i.e. binary classification task). To do this, we seek to match $S$ to a bank of $\ell$ learned memory matrices, where each matrix is loosely representative of a motif. Our output classifier is then based on a linear combination of the matches.

Each memory bank "motif" $M_i$ is a learned matrix $\in \mathbb{R}^{p \times t}$, and there are $\ell$ total motif matrices. We embed the input matrix $S$ and memory matrices $\{M_1, ..., M_\ell\}$ with two separate functions, $f$ and $g$, respectively. $f$ is a bidirectional attention LSTM on $S$, where we compute a context vector $s \in \mathbb{R}^d$ based on the attention weights $\alpha_i$ as in Bahdanau et al. (2014). $g$ is similar to the $g$ function in Vinyals et al. (2016). We first embed each memory matrix $M_i$ using the context vector of a bidirectional attention LSTM $g'(M_i)$, and then pass all of the $g'()$ outputs through another bidirectional LSTM $g()$ to obtain a relationship among the memory motifs, producing final memory embeddings $\{m_1, ..., m_\ell\} \in \mathbb{R}^d$:

$$\begin{aligned} s &= f(S) \\ m_i &= g(g'(M_i)) \text{ for } i \text{ in } 1, ..., \ell \end{aligned} \quad (1)$$

Once we have the embedding vectors, we want to compare the original input sequence to the memory units for classification. For the current $S$, we first we want to find the importance, or weight $w_i$, of each memory matrix $M_i$. We do this by using a normalized cosine similarity score between $s$ and each $\{m_1, ..., m_\ell\}$, which is the matching part of the network:

$$w_i = \frac{\exp\left(K[s, m_i]\right)}{\sum_j \exp\left(K[s, m_j]\right)} \quad (2)$$

where $K$ is cosine similarity. We then read from the memory embeddings using the matching weights to produce a read vector $r$, which is a linear combination of the memory embeddings multiplied by their respective matching weights. This allows for the sequence to not have to match to a particular motif, but rather a combination of them:

$$r = \sum_i w_i \mathbf{m}_i \quad (3)$$

The read vector and original input embedding vector are then linearly transformed, added together, and then fed through a softmax equation to get $\mathbf{y}$, the probability of $X$ being a positive TFBS:

$$\begin{aligned} \tilde{\mathbf{y}} &= \mathbf{W}^s s + \mathbf{W}^r r \\ \mathbf{y} &= softmax(\tilde{\mathbf{y}}) \end{aligned} \quad (4)$$

Table 1: TFBS Binary Classification Results

| Model | Mean AUC | Median AUC | Stdev |
|---|---|---|---|
| PWM | 0.850 | 0.876 | 0.120 |
| LSTM | 0.910 | 0.932 | 0.056 |
| CNN | 0.918 | 0.940 | 0.084 |
| MMN (ours) | 0.926 | 0.950 | 0.068 |

## 3 EXPERIMENTS

We ran our MMN model on the 61 leukemia cell TF datasets which had a training set of at least 10,000 sequences from Alipanahi et al. (2015). Each TF dataset has exactly 1,000 testing sequences. All training and testing sets have an even positive/negative TFBS sequence split. Each sequence sample $S$ is 101 length ($t = 101$) and composed of DNA-base characters (A,C,G,T). Since there is a separate dataset for each different TF, we train a separate model for each TF. In other words, each model constructs its own memory bank for that particular TF. We then aggregate the accuracy results over all TFs for model comparison. In our experiments, we tuned the following hyperparameters: number of memory matrices $\ell \in \{2, 4, \mathbf{8}, 16\}$, lookup table vector size $p \in \{2, \mathbf{4}, 8, 16\}$, embedding vector size $d \in \{32, 64, \mathbf{128}\}$, with the best parameters in bold.

### 3.1 ACCURACY RESULTS

We compare our method to the baseline PWM-motif matching approach from Machanick & Bailey (2011), as well as the CNN model in Alipanahi et al. (2015), and a regular LSTM model in Lanchantin et al. (2016). The results are shown in Table 1. Our model significantly outperforms the baseline of manually constructed PWMs (based on a pairwise t-test). Our model performs slightly better than the other two baseline deep learning models, including the CNN, which should be able to extract motifs automatically via filters. We believe that it is important to show that memory matching models are powerful for classification.

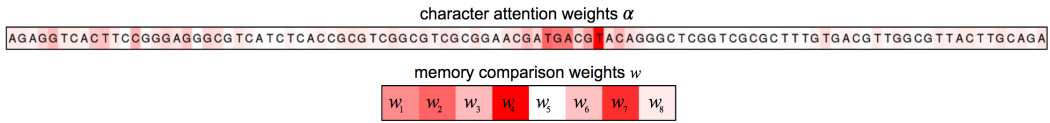

Figure 2: Visualizing the Model: Character and memory weights for a positive ATF1 binding sequence

### 3.2 VISUALIZING AND INTERPRETING THE MODEL

In medical classification tasks, if a model is accurate, yet hard to interpret, doctors may be reluctant to use it. Thus, we seek to understand our model via visualization. While it is difficult to correlate the memory matrices with DNA characters since they are in an embedded space, we can look at individual test sequences and see what the model focuses on. We do this by viewing the attention weights $\alpha_i$ from $f$ to see which characters are most important, and also the memory matching weights $w_i$ to see which memory matrices closely match this sequence. Figure 2 shows an example of our visualization technique where the sequence is a positive TFBS sequence from the "ATF1" TF. The character $\alpha$ attention mostly focuses on the subsequence *TGACGTA* in the middle, which is a "known" motif for ATF1 (Mathelier et al., 2016). Seeing which particular memory units match the test sequence via $w_i$ doesn't tell us anything, but it may be of use for comparing multiple sequences to see if they match certain memories. We plan to explore the interpretability of the learned memory in future work.

## 4 CONCLUSION

In this work, we introduced memory matching networks (MMNs) for learning a set of memory matrices which can be matched to a new test sequence for classification. We applied our model on the binary classification task of predicting the binding sites of Transcription Factor proteins, where a memory of learned "motifs" is beneficial for classifying sequences. We showed that it outperforms the baseline models of PWM-matching and a CNN, as well as a way to visualize the predictions. We hope that this work will provoke research on memory-based models for genomic sequences, where it is a fitting setting. Although we use this on a medical task, we think that memory matching networks may be of use in any application where classification can be done by matching to memory such as text or image classification.